\documentclass{article}

\usepackage[preprint]{neurips_2024}

\usepackage[utf8]{inputenc} 
\usepackage[T1]{fontenc}    
\usepackage{hyperref}       
\usepackage{url}            
\usepackage{booktabs}       
\usepackage{amsfonts}       
\usepackage{nicefrac}       
\usepackage{microtype}      
\usepackage{xcolor}         

\newcommand{\name}{EXAONEPath}

\newcommand{\blue}[1]{\textcolor{blue}{#1}}

\usepackage{kotex}
\usepackage{booktabs} 
\usepackage{subcaption}
\usepackage{wrapfig}
\usepackage{enumitem}
\usepackage[hang, flushmargin]{footmisc}
\usepackage{hyperref}

\usepackage{multirow}
\usepackage{graphicx}
\usepackage[normalem]{ulem} 
\useunder{\uline}{\ul}{} 
\usepackage{adjustbox}
\usepackage{array}

\usepackage{natbib}   
\title{EXAONEPath 1.0 Patch-level Foundation Model for Pathology}

\author{Juseung Yun
,Yi Hu, Jinhyung Kim, Jongseong Jang \& Soonyoung Lee 
\\
LG AI Research\\
Seoul, South Korea \\
\texttt{\{js.yun
Yi Hu,Jinhyung.Kim,j.jang,soonyoung.lee\}@lgresearch.ai} \\
}

\begin{document}

\maketitle

\begin{abstract}
Recent advancements in digital pathology have led to the development of numerous foundational models that utilize self-supervised learning on patches extracted from gigapixel whole slide images (WSIs).
While this approach leverages vast amounts of unlabeled data, we have discovered a significant issue: features extracted from these self-supervised models tend to cluster by individual WSIs, a phenomenon we term WSI-specific feature collapse.
This problem can potentially limit the model's generalization ability and performance on various downstream tasks.
To address this issue, we introduce \name{}\footnote{github:\href{https://github.com/LG-AI-EXAONE/EXAONEPath}{https://github.com/LG-AI-EXAONE/EXAONEPath}\\ huggingface:\href{https://huggingface.co/LGAI-EXAONE/EXAONEPath}{https://huggingface.co/LGAI-EXAONE/EXAONEPath}}
, a novel foundational model trained on patches that have undergone stain normalization. 
Stain normalization helps reduce color variability arising from different laboratories and scanners, enabling the model to learn more consistent features. 
\name{} is trained using 285,153,903 patches extracted from a total of 34,795 WSIs. 
Our experiments demonstrate that \name{} significantly mitigates the feature collapse problem, indicating that the model has learned more generalized features rather than overfitting to individual WSI characteristics. 
We compared \name{} with state-of-the-art models across six downstream task datasets, and our results show that \name{} achieves superior performance relative to the number of WSIs used and the model's parameter count. 
This suggests that the application of stain normalization has substantially improved the model's efficiency and generalization capabilities. 
\end{abstract}

\section*{\blue{*Important Notice*}}
\blue{This document is valid only for \name{} version 1.0. Also, please refer to the license at the end of the document, when using this model.}

\section{Introduction}
In modern pathology, the analysis of digital pathology images plays a crucial role in cancer subtyping \citep{clam,transmil}, prognosis prediction \citep{prognosis_prediction1,prognosis_prediction2,prognosis_prediction3}, and quantifying the tissue microenvironment \citep{microenvironment1,microenvironment2,microenvironment3,hovernet}.
Whole Slide Images (WSIs), in particular, are high-resolution digitized tissue slides that often measure tens of thousands by tens of thousands of pixels in size. 
Due to the vast dimensions of these WSIs, many researchers have adopted the approach of dividing the images into smaller patches to train their models. 
A single WSI can be divided into thousands or even tens of thousands of patches, but accurately labeling each of these patches is a significant burden in terms of time and cost. 
To overcome this limitation, recent approaches have widely adopted self-supervised learning techniques to first train a foundation model, which is then fine-tuned for various downstream tasks \citep{lunit,UNI,virchow,rudolfv}. 
This approach allows for the effective utilization of large amounts of unlabeled data while achieving high performance with limited labeled data.

In this paper, we first introduce our novel observation that the features of models trained through self-supervised learning exhibit an unexpectedly high degree of overlap.
Specifically, we divide whole slide images into patches and create a foundation model using the DINO (self-DIstillation with NO labels) \citep{dino} self-supervised learning method.
During this process, as the whole slides are divided into small patches for training, the model has no information about which whole slide each specific patch originated from.
Subsequently, we extract features of each patch and visualize them using t-SNE \citep{t-sne}.
Surprisingly, we observe that the patch features cluster according to their respective WSIs. 
We refer to this phenomenon as WSI-specific feature collapse. 
Whole slides undergo Hematoxylin and Eosin (H\&E) staining to clearly reveal tissue structures. 
We hypothesize that this clustering phenomenon occurs due to variations in staining intensity between laboratories or even within the same laboratory, resulting in color differences among WSIs.
In whole slide image analysis, model features should focus on capturing pathologically significant information rather than simply relying on the degree of staining.
Critical information that should be observed includes nuclear size and shape, cell density, and structural changes in tissue. 
If the model concentrates excessively on staining variations, it may overlook these essential pathological features, potentially leading to inaccuracies in important downstream tasks such as cancer type classification, mutation classification, and survival analysis. 
Therefore, it is crucial to ensure that the model is trained to properly recognize and learn these important pathological characteristics during the training process.
To address this issue, we first apply Macenko normalization \citep{macenko} to the patches before feature extraction in an effort to standardize the color characteristics across patches.
While this somewhat mitigates the feature collapse, significant collapse is still observed.
To further address this challenge, we introduce \name{}, a stain-normalized pathology foundation model. 
This model is created by performing DINO self-supervised learning pretraining on patches that have undergone Macenko normalization.
\name{} is trained using 285,153,903 patches extracted from a total of 34,795 WSIs. 
We have confirmed that \name{} significantly mitigates the feature collapse problem. 
This implies that the model has a higher potential to extract pathologically relevant features crucial for downstream tasks, rather than simply relying on color-based characteristics. 
These improvements are expected to enhance the model's performance and reliability across various pathological analysis tasks.
We validated the performance of \name{} on various patch-level tasks including PCAM \citep{pcam1,pcam2}, MHIST \citep{mhist}, CRC 100k \citep{crc100k}, TIL-Detection \citep{til_det1,til_det2,til_det3}, MSI-CRC, and MSI-STAD \citep{msi_crc_stad} dataset. 
Notably, despite being trained on relatively less data compared to existing State-of-the-Art (SOTA) models, and using only publicly available datasets, \name{} achieved comparable performance levels to these SOTA models across these diverse tasks.
Our contributions can be summarized as follows:


\begin{figure}[t!]
    \centering
    \begin{subfigure}[t]{0.48\textwidth}
        \centering
        \includegraphics[width=\textwidth]{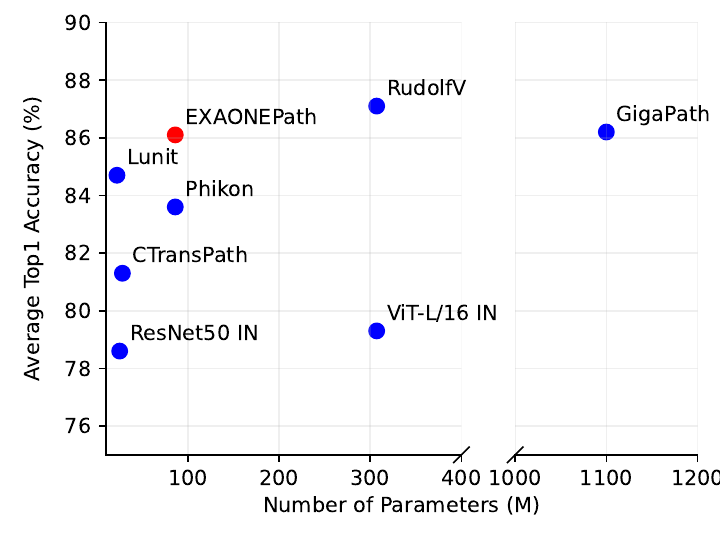}
        \caption{}
        \label{fig:param_comparison}
    \end{subfigure}
    \hfill
    \begin{subfigure}[t]{0.48\textwidth}
        \centering
        \includegraphics[width=\textwidth]{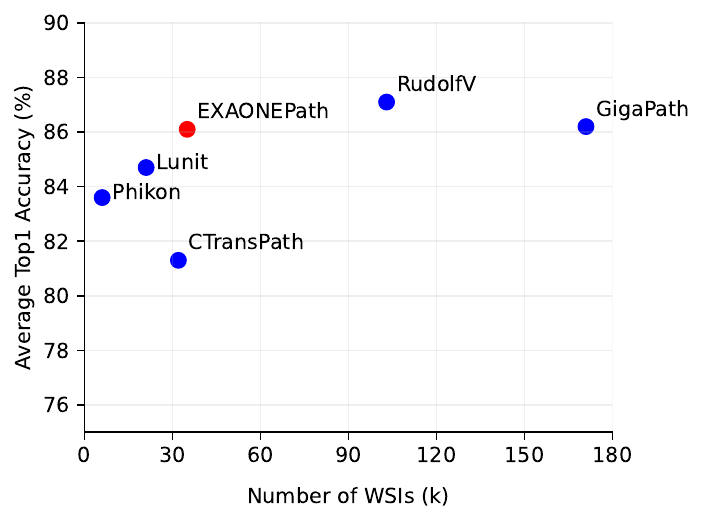}
        \caption{}
        \label{fig:wsis_comparison}
    \end{subfigure}
    \caption{\textbf{Performance comparison of models based on the number of parameters and the number of WSIs used for training.} The average Top-1 accuracy represents the mean linear evaluation performance across six downstream tasks.
(a) Average Top-1 accuracy versus the number of parameters. (b) Average Top-1 accuracy versus the number of WSIs used for training. Notably, our model (\name{}) achieves high performance despite having fewer parameters and using fewer WSIs compared to other models, demonstrating its efficiency.}
    \label{fig:model_comparison}
\end{figure}


\begin{itemize}
    \item We discover and analyze the ‘WSI-specific feature collapse' phenomenon, where patch features cluster according to their source WSI, despite the foundation model being trained without information about which WSI each patch was extracted from.
    \item To address this issue, we develop \name{}, a novel model that applies stain normalization to patches during foundation model training. We demonstrate that \name{} effectively mitigates the feature collapse problem.
    \item We evaluate \name{} against state-of-the-art models on six diverse patch-level tasks. The results show that \name{} exhibits superior performance relative to the number of WSIs used and the number of model parameters.
\end{itemize}

\section{Related Work}
In this section, we briefly overview some relevant literature. 




\textbf{Pathology Image Foundation Models.}
With the significant advancements in self-supervised learning (SSL) methods, there is a notable trend in applying these methods, commonly used in natural image processing, to histopathology to develop robust foundation models for whole slide images (WSIs) \citep{lunit_ref1,lunit_ref2,lunit_ref3,lunit}. Given the gigapixel resolution of WSIs, creating foundation models by dividing images into patch units has become a prevalent approach.
For example, REMEDIS \citep{REMEDIS} employs SimCLR \citep{simclr} to train on various medical domain images, including pathology. Similarly, \citet{lunit_ref2} utilizes SimCLR for training on histopathology images.
\citet{lunit} applies four SSL methods—MoCo v2 \citep{mocov2}, SwAV \citep{swav}, Barlow Twins \citep{barlow_twins}, and DINO \citep{dino}—to pathology data and evaluates their performance on image classification and nuclei instance segmentation tasks.
\citet{virchow_ref1} compares the Masked Autoencoder (MAE) \citep{mae} and DINO \citep{dino} algorithms using the largest pathology dataset to date, which comprises over three billion images. Their evaluation across six clinical tasks demonstrates that pre-training on pathology data is more effective than pre-training on natural images, with the DINO algorithm showing superior generalization performance across all tasks.
HIPT \citep{HIPT} also implements DINO within a hierarchical structure.
Phikon \citep{phikon} employs iBOT \citep{ibot}, a self-supervised learning method based on masked image modeling (MIM), for histology images. The ViT-Base model using iBOT exhibits excellent performance across 17 downstream tasks on 7 cancer types, suggesting its potential for developing a foundation model for histopathology.
CTransPath \citep{CTransPath} introduces a new self-supervised learning strategy called Semantically-Relevant Contrastive Learning (SRCL), an enhancement of MoCo v3 \citep{mocov3}.
Additionally, UNI \citep{UNI}, RudolfV \citep{rudolfv}, Virchow \citep{virchow}, and Prov-GigaPath \citep{gigapath} utilize DINO v2 for training pathology foundation models. Prov-GigaPath \citep{gigapath} applies the LongNet \citep{longnet} architecture to learn slide-level representations.

To our knowledge, none of these studies utilizing self-supervised learning for developing pathology foundation models employ stain normalization on the training data. In contrast, we incorporate stain normalization in our training process, which we believe offers significant advantages for learning robust foundation models.

\textbf{Stain Normalization.}
Stain normalization is a critical process in histopathology designed to address the inconsistencies in color that arise from staining protocols, device properties, scanner settings, and tissue preparation methods.
These variations can affect the accuracy of automated diagnostic systems. Therefore, robust methods to standardize the appearance of histopathological images are essential to ensure reliable and reproducible analyses.

Early methods focused on histogram transformation, such as Reinhard normalization \citep{reinhard}, which employs statistical techniques to transfer the color characteristics from one image to another by converting them to the Lab color space.
This method is widely recognized for its simplicity and effectiveness in standardizing color distributions across different images. 
Another notable approach is Macenko normalization \citep{macenko}, which transforms images into an optical density (OD) space. PCA is then performed on the OD space to identify the principal components corresponding to the Hematoxylin and Eosin stains, thus enabling effective stain separation and normalization.
Inspired by the biological fact about stain binding, \citet{Vahadane} extended the use of nonnegative matrix factorization (NMF) \citep{NMFstain} by introducing sparsity constraints, which enhance the robustness of stain separation by limiting the number of non-zero components, thereby aligning with the actual biological composition of histological samples.

Recently, deep learning-based approaches have advanced stain normalization techniques, particularly through the use of image-to-image translation architectures combined with adversarial learning methods \citep{StainTransfer}. 
These approaches leverage generative adversarial networks (GANs) \citep{goodfellow2014generative}, such as CycleGAN \citep{cyclegan} and StarGAN \citep{choi2018stargan}, to learn the mappings between different staining styles.
By employing adversarial training, these models are able to produce normalized images that closely resemble the target staining characteristics while preserving essential structural features of the tissue \citep{HistAuGAN, MultiPathGAN}. 
Such advancements demonstrate the potential of deep learning to overcome the limitations of traditional stain normalization methods, offering more robust and scalable solutions for digital pathology.

\section{Self-supervised Pre-training for Pathology}
\begin{figure}[tbp]
    \centering
    \begin{subfigure}[t]{0.32\textwidth}
        \centering
        \includegraphics[width=\textwidth,height=0.8\textheight,keepaspectratio]{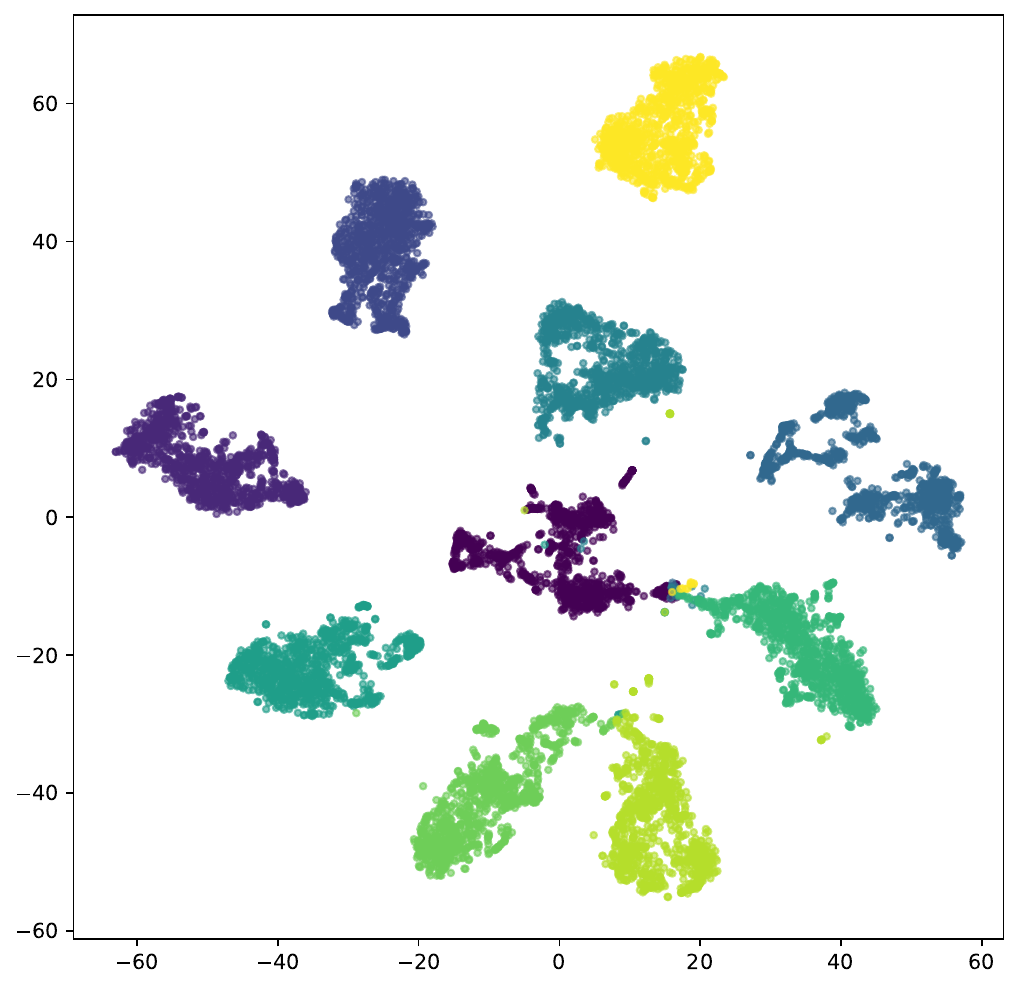}
        \caption{}
        \label{fig:features_dino}
    \end{subfigure}%
    \hfill%
    \begin{subfigure}[t]{0.32\textwidth}
        \centering
        \includegraphics[width=\textwidth,height=0.8\textheight,keepaspectratio]{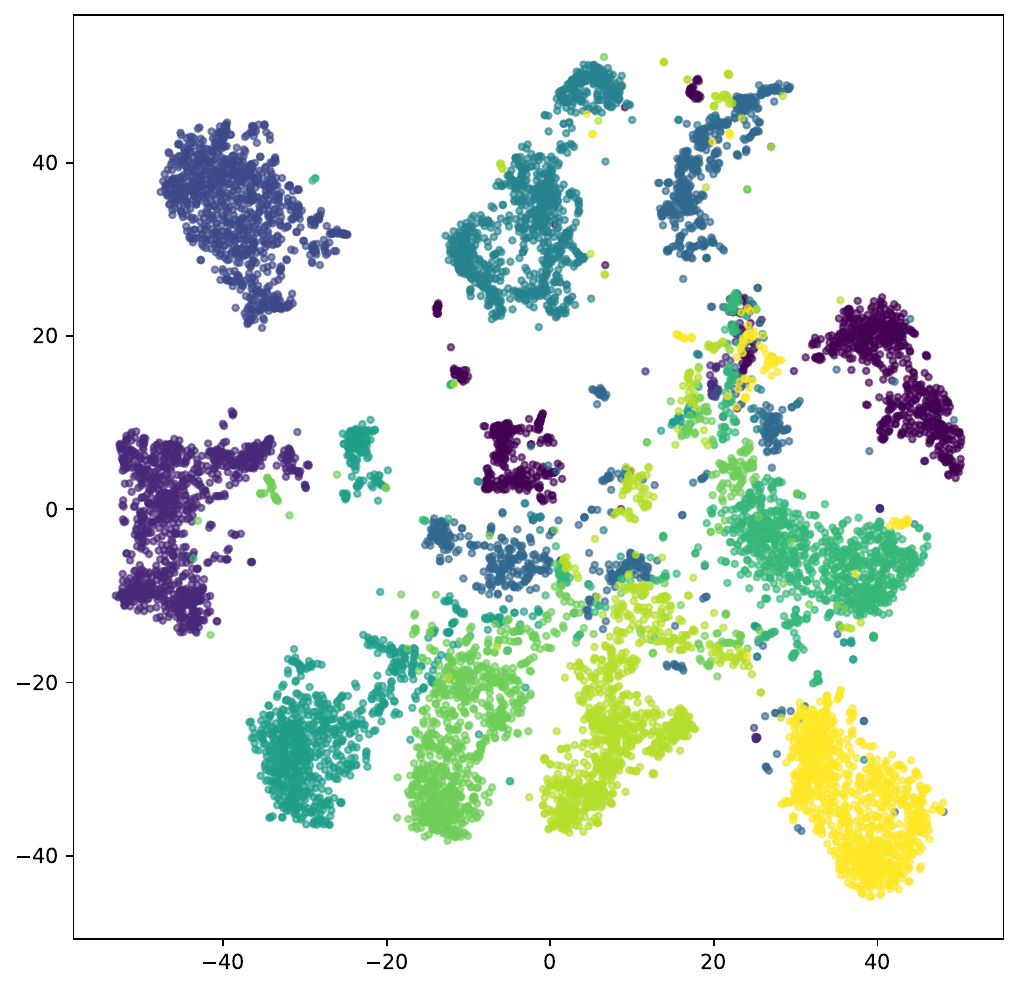}
        \caption{}
        \label{fig:features_macenko}
    \end{subfigure}%
    \hfill%
    \begin{subfigure}[t]{0.32\textwidth}
        \centering
        \includegraphics[width=\textwidth,height=0.8\textheight,keepaspectratio]{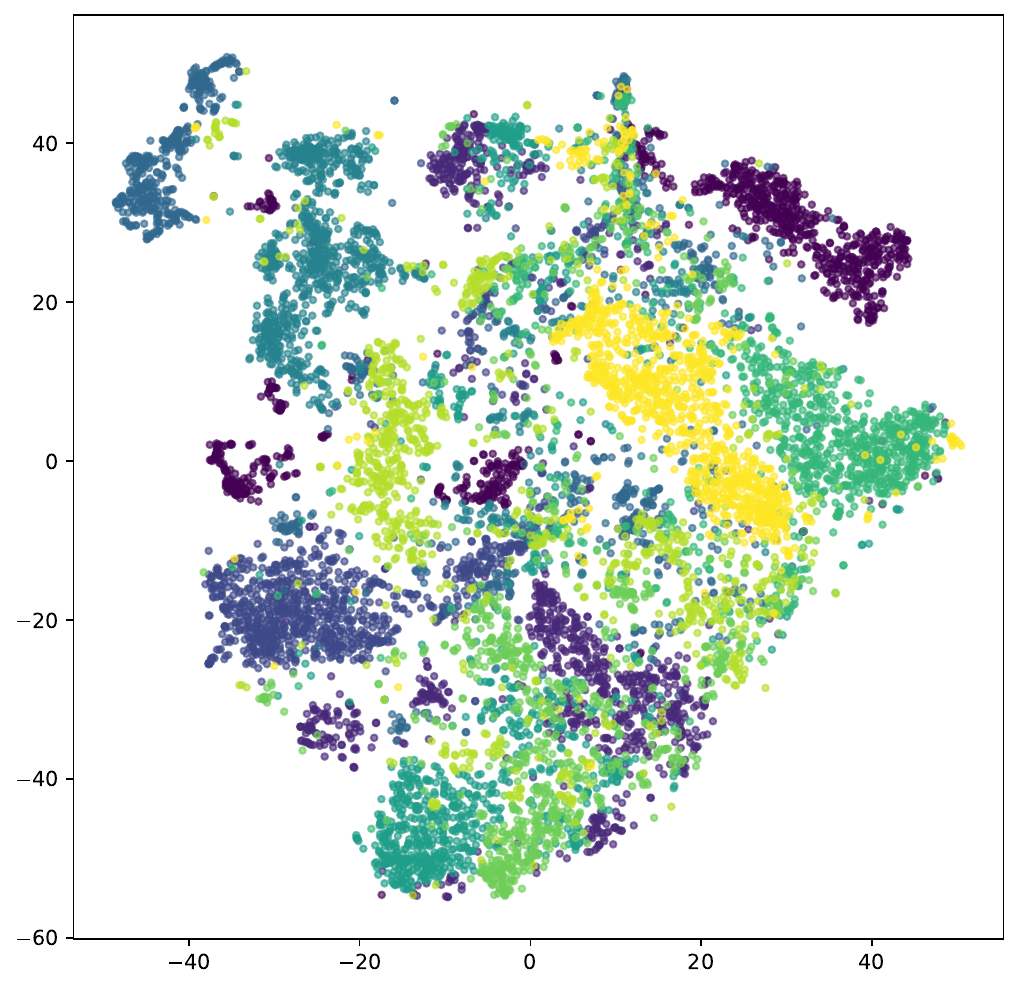}
        \caption{}
        \label{fig:features_dino_macenko}
    \end{subfigure}
    \caption{\textbf{t-SNE visualization of features extracted from a foundation model trained with DINO}. 1000 patches are randomly sampled from each of 10 arbitrarily selected WSIs, and the features from each of the 10 WSIs are represented in different colors. Despite having no information about the source WSI for each patch, the model exhibits WSI-specific feature collapse, where features tend to cluster by WSI.
    (a) Features obtained by inputting patches without any stain normalization into a model trained on non-stain-normalized images, showing severe feature collapse.
    (b) Features obtained by inputting stain-normalized patches into a model trained on non-stain-normalized images, showing significant but reduced collapse compared to (a).
    (c) Features obtained by inputting stain-normalized patches into a model trained on stain-normalized images, showing significantly reduced feature collapse, as proposed by our method.}

    \label{fig:feature_collapse}
\end{figure}
This section introduces the ‘WSI-specific feature collapse’ that occurs when pathology foundation models are trained using self-supervised learning, and discusses methods to reduce this phenomenon.
In Section \ref{sec:feature collapse}, we discuss the 'WSI-specific feature collapse' problem.
This refers to the phenomenon where features from the same WSI tend to cluster together, even though each patch is learned without information about which WSI it comes from during the training process.
Section \ref{sec:stain normalization} demonstrates that applying stain normalization before feature extraction does not significantly reduce the collapse.
In Section \ref{sec:stain_normalized_model}, we introduce \name{}, which applies stain normalization to the training data from the beginning of foundation model training to further reduce ‘WSI-specific feature collapse'.

\subsection{motivation: wsi specific feature collapse}
\label{sec:feature collapse}
DINO \citep{dino} involves applying different augmentations to an image and inputting them into both teacher and student models, training them to produce identical features. 
Color jittering is a data augmentation technique commonly used in self-supervised learning, including DINO. 
It involves randomly altering the colors of an image to create different variations, helping the model learn robust features invariant to color changes.
However, despite the use of color jittering augmentation, when visualizing patch features using t-SNE \citep{t-sne}, we observe that features still tend to collapse according to their source WSI.
Specifically, we trained a foundation model using DINO on $256^2$-sized patches extracted from gigapixel WSIs.
It is important to note that the neural network is trained only on these isolated patches, without any information about which WSI each patch originated from. 
We randomly selected 1,000 patches from each of 10 arbitrary WSIs, performed a forward pass through the trained foundation model, and then visualized the resulting 10,000 features using t-SNE (see Figure \ref{fig:features_dino}).
Interestingly, even though the model does not know the source WSI of each patch, we observe that patches from the same WSI tend to have similar features and cluster together.
This phenomenon suggests that, instead of learning features crucial for downstream tasks and pathologically significant, the model may have unintentionally learned WSI-specific characteristics, such as staining intensity or subtle differences due to scanner variations.
This feature collapse phenomenon could potentially compromise the model's ability to generalize.
If the model becomes overly adapted to individual WSI characteristics, its performance may degrade when applied to new WSIs or images acquired from different institutions.
This result suggests that when creating a pretrained foundation model for WSIs, rather than directly applying self-supervised learning methods used for natural images, we need to use approaches that take into account the specific characteristics of WSIs.

\subsection{stain normalization}
\label{sec:stain normalization}
We hypothesized that the WSI-specific feature collapse occurs due to variations in staining intensity across different WSIs, which result in color differences.
This phenomenon arises from the staining process of WSIs, which depends on the types of stains used and the degree of staining.
Additionally, differences in slide scanner settings or performance can affect color variations in digitized WSIs.
Therefore, to address these color differences, we employed Macenko stain normalization \citep{macenko} before extracting features. 
We used the same pretrained model and patches as described in Section \ref{sec:feature collapse}.
Figure \ref{fig:features_macenko} shows the results. 
Compared to Figure \ref{fig:features_dino}, where features are completely separated by WSI, Figure \ref{fig:features_macenko} shows that features from different WSIs are now slightly mixed, but they still tend to cluster by WSI.
This indicates that while Macenko normalization partially mitigates color differences between WSIs, it is not sufficient to fully resolve the issue, as features from the same WSI still exhibit noticeable clustering and the collapse problem remains significant.
Furthermore, it suggests that applying Macenko normalization only during inference has limited effectiveness because the pretrained model has already learned to extract features biased toward color characteristics.

\subsection{stain normalized whole slide pathology image foundation model}
\label{sec:stain_normalized_model}
We use DINO \citep{dino} self-supervised learning to train a WSI foundation model.
We refer to this model as \name{}, which is trained with stain-normalized data using DINO.
DINO involves applying different augmentations to an image and inputting them into both teacher and student models, training them to produce identical features. 
The teacher model, updated with an exponential moving average (EMA) of the student parameters, guides the output of the student model.
The student model is trained using cross-entropy loss to measure their similarity.
By inputting differently augmented images into the teacher and student models, the features produced by both models are compared and aligned. 
In the original DINO paper, different augmentation methods are applied to two $256^2$ global views and several $96^2$ local views of each image.

Combining Macenko normalization with the existing DINO augmentation is a design choice. 
We apply Macenko normalization to all images with 100\% probability before applying DINO augmentation.
This approach is chosen because applying Macenko normalization to each global view and local view individually is computationally intensive, creating a bottleneck in the data loader during training.
Figure \ref{fig:features_dino_macenko} visualizes the features of \name{} trained with DINO using Macenko-normalized images.
Macenko normalization is applied not only during training but also when extracting features for each patch in the visualization process. 
Compared to Figures \ref{fig:features_dino} and \ref{fig:features_macenko}, we observe that patch features from various WSIs are much more intermingled.
This indicates that the model learns more generalized features and is less dependent on the source of the WSIs.
This allows the model to focus on more histologically relevant features.
We still observe that some clusters of features are grouped by WSI. We believe this occurs for two reasons: first, while Macenko normalization significantly reduces color differences, it may not completely eliminate them; second, patches from the same WSI likely share clinically relevant features beyond just color, due to variations in tissue type, cancer subtype, and patient characteristics within each WSI.

\section{Experiments}
In this section, we describe the experimental settings and results.
Section \ref{sec:pretraining data} provides a detailed explanation of the data used for pretraining the whole slide image foundation model.
Section \ref{sec:pretraining details} offers comprehensive information on the pretraining methodology.
Section \ref{sec:downstream task data} describes the data for the downstream tasks we aim to evaluate.
Section \ref{sec:downstream task training details} explains in detail the methods used to training the downstream tasks.
Finally, Section \ref{sec:downstream task results} presents the results of the linear evaluation on the downstream tasks.

\subsection{pretraining data}
\label{sec:pretraining data}
\begin{figure}[tbp]
    \centering
    \includegraphics[width=0.45\textwidth,height=0.8\textheight,keepaspectratio]{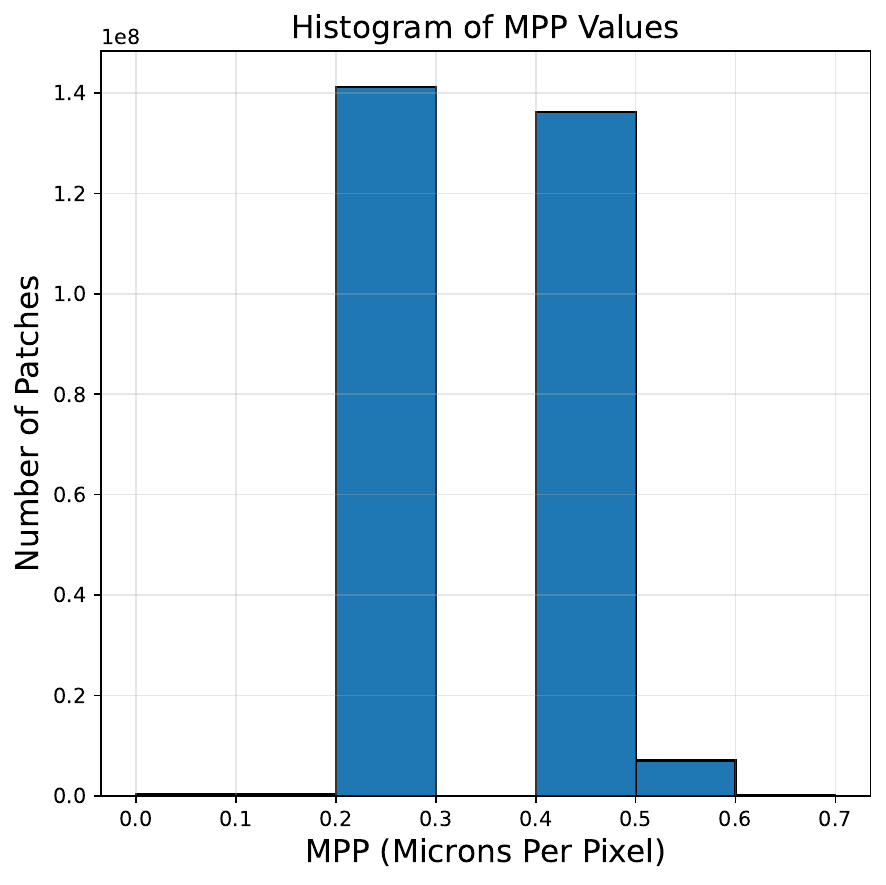}
    \caption{\textbf{MPP distribution of the training data}. Most of the data used for training is concentrated around 0.25 MPP and 0.5 MPP.}
    \label{fig:mpp_distribution}
\end{figure}
We collected 34,795 Formalin-Fixed, Paraffin-Embedded (FFPE) Hematoxylin and Eosin (H\&E) stained whole slide images (WSIs).
Following CLAM \citep{clam}, we divide the WSIs into non-overlapping patches, focusing only on the tissue-containing regions. 
Given that each WSI has a different micron per pixel (MPP), we adjust the patch sizes so that they correspond to 0.5 MPP when resized to $256^2$. 
For instance, for WSIs with 0.25 MPP, we extract patches at $512^2$ dimensions, and for WSIs with 0.5 MPP, we extract patches at $256^2$ dimensions. 
However, for training the \name{} model, the patches are not resized to achieve a uniform 0.5 MPP. 
Instead, we retain the original patch sizes during training because the DINO algorithm's input image transform includes random resized cropping. 
Even if we resize the patches to 0.5 MPP, the model's input would still undergo changes in MPP due to this random resized crop operation. 
Therefore, maintaining the originally extracted patch sizes allows us to preserve the maximum amount of information from each WSI while relying on DINO's augmentations to introduce the necessary variability in scale and perspective. 
Figure \ref{fig:mpp_distribution} presents the distribution of MPP values for patches extracted from the WSIs. 
Most of the data used for training is concentrated around 0.25 MPP and 0.5 MPP. 
The total number of patches used for training \name{} is 285,153,903.

\subsection{pretraining details}
\label{sec:pretraining details}
We train \name{} using DINO \citep{dino} with a ViT-B model \citep{vit}, which has a patch size of 16, initializing the weights with those pretrained on the ImageNet dataset \citep{imagenet}.
Training is conducted on 10 machines, each equipped with 8 A100 GPUs. 
We use a batch size of 5,120 and a learning rate of 0.005, training the model for 10 epochs. 
The learning rate is warmed up over the first 1,000 iterations. 
We employ bfloat16 precision training. 
The output layer is not fixed; instead, it is trained from scratch. 
Additionally, we set the number of local crops to 8.
Other hyperparameters follow the settings from the original DINO paper for training the ViT-B model.

\subsection{downstream task data}
\label{sec:downstream task data}
We evaluate \name{} on six publicly available patch-level datasets for classification tasks: PCAM \citep{pcam1,pcam2}, MHIST \citep{mhist}, CRC-100K \citep{crc100k}, TIL Detection \citep{til_det1,til_det2,til_det3}, MSI CRC \citep{msi_crc_stad}, and MSI STAD \citep{msi_crc_stad}.

\textbf{PCAM} \citep{pcam1,pcam2} consists of 327,680 color images, each of $96^2$ pixels, extracted from histopathologic scans of lymph node sections, and each annotated with a binary label indicating the presence of metastatic tissue.
The dataset is divided into a training set of 262,144 examples and validation and test sets of 32,768 examples each, with no overlap between WSIs across splits.
All splits maintain a 50/50 balance between positive and negative examples. 
A positive label indicates the presence of at least one pixel of tumor tissue within the central $32^2$ pixel region of a patch, while tumor tissue in the outer region does not affect the label. 
PCAM is derived from the Camelyon16 Challenge, which includes 400 H\&E-stained WSIs of sentinel lymph node sections, digitized at 40x magnification (0.243 MPP) and undersampled to 10x for a larger field of view.
The dataset follows the train/test split from Camelyon16, with 20\% of the training WSIs held out for validation. 

\textbf{MHIST} \citep{mhist} consists of 3,152 H\&E-stained FFPE images, each $224^2$ pixels in size, of colorectal polyps, collected from the Department of Pathology and Laboratory Medicine at Dartmouth-Hitchcock Medical Center (DHMC).
Each image is scanned at 40x magnification and resized to 8x magnification and is labeled based on the consensus of seven pathologists at DHMC, categorizing the type of colorectal polyps.
The MHIST dataset focuses on a binary classification task, distinguishing between Hyperplastic Polyps (HP) and Sessile Serrated Adenomas (SSA).
The dataset is divided into training and test sets, with the training set further split into training and validation sets using an 80:20 ratio. 
This results in 1,740 training samples, 435 validation samples, and 977 test samples.

\textbf{CRC-100K} \citep{crc100k} consists of 100,000 non-overlapping image patches ($224^2$ pixels, 0.5 MPP) from H\&E-stained histological images of human colorectal cancer (CRC) and normal tissue. 
The dataset includes both color-normalized images using Macenko's method, and non-normalized images, with the latter exhibiting slight variations in staining intensity and color. 
We utilize the non-normalized images for our work, as we intend to apply Macenko normalization using our Macenko target image.
Tissue classes include Adipose (ADI), Background (BACK), Debris (DEB), Lymphocytes (LYM), Mucus (MUC), Smooth Muscle (MUS), Normal Colon Mucosa (NORM), Cancer-associated Stroma (STR), and Colorectal Adenocarcinoma Epithelium (TUM). 
These patches are manually extracted from 86 H\&E-stained FFPE samples from the NCT Biobank and the UMM pathology archive. 
An additional set, CRC-VAL-HE-7K, consists of 7,180 patches from 50 patients with CRC and serves as a validation set with no overlap with the original dataset. 
Since there is no separate test set provided, we use CRC-VAL-HE-7K as our test set. 
For training and validation, we split the original training set into an 80:20 ratio to create training and validation sets.

\textbf{TIL Detection} \citep{til_det1,til_det2,til_det3} dataset consists of 304,097 H\&E-stained images, each $100^2$ pixels at 0.5 MPP, extracted from FFPE sections covering 23 different cancer types from TCGA \citep{tcga}.
The dataset is used for predicting TIL-positive images, where an image is considered positive if it contains more than two tumor-infiltrating lymphocytes (TILs). 
The dataset is divided into 209,221 training images, 38,601 validation images, and 56,275 test images. 
Each image is annotated as either TIL-positive (54,910 images) or TIL-negative (249,187 images), based on the presence of at least two TILs.

\textbf{MSI CRC} and \textbf{MSI STAD} \citep{msi_crc_stad}. For MSI CRC, the training set contains 93,408 image patches, and the test set comprises 99,904 image patches. 
For MSI STAD, the training set includes 100,570 image patches, and the test set contains 118,008 image patches. 
All images are $224^2$ pixels at 0.5 MPP and are derived from histological images of colorectal and gastric cancer patients in the TCGA cohort. 
The images are sourced from FFPE diagnostic slides.
Preprocessing includes automatic tumor detection, resizing to $224^2$ pixels, and color normalization using Macenko's method.
Patients are categorized into "MSS" (microsatellite stable) or "MSIMUT" (microsatellite instable or highly mutated) groups, and the data is split into training (70\%) and testing (30\%) sets at the patient level, with the training set balanced by undersampling excess MSS tiles.
We further split the training set into training and validation sets with an 80:20 ratio.

\subsection{Downstream Task Training Details}
\label{sec:downstream task training details}
We use linear evaluation, a common practice in previous SSL research for evaluation \citep{simclr,barlow_twins,moco}. 
Training is performed using the SGD optimizer with a learning rate of 0.1, without weight decay, and a momentum of 0.9, with a batch size of 128 for 12,500 iterations. 
The images are first resized to $256^2$ pixels and then center-cropped to $224^2$ pixels. 
No additional data augmentation is applied. 
Macenko normalization is used in both the training and evaluation processes. 
Even datasets that are already Macenko-normalized are re-normalized using the same Macenko target images employed during \name{} training.

\subsection{Downstream Task Results}
\label{sec:downstream task results}
\begin{table}[!t]
\centering
\caption{\textbf{Linear evaluation performance on six downstream tasks.} \name{} is compared with other models, including state-of-the-art pathology foundation models. 
Top-1 accuracy is shown. Values for models other than GigaPath are taken from the RudolfV paper}
\label{tab:linear eval result}
\resizebox{\textwidth}{!}{%
\begin{tabular}{@{}lrrrccccccc@{}}
\toprule
\multicolumn{1}{c}{} &
  \multicolumn{2}{c}{\# data} &
  \multicolumn{1}{c}{} &
  \multicolumn{7}{c}{Dataset} \\
\cmidrule(lr){2-3} \cmidrule(lr){5-11}
\multicolumn{1}{c}{\multirow{-2}{*}{Model}} &
  \multicolumn{1}{c}{WSI} &
  \multicolumn{1}{c}{Patch} &
  \multicolumn{1}{c}{\multirow{-2}{*}{\# of params}} &
  PCAM &
  MHIST &
  CRC-100K &
  TIL Det. &
  MSI CRC &
  MSI STAD &
  Avg \\ \midrule
ResNet50 ImageNet &
  \multicolumn{1}{c}{-} &
  \multicolumn{1}{c}{-} &
  &
  0.833 &
  0.806 &
  0.849 &
  0.915 &
  0.653 &
  0.664 &
  0.786 \\
ViT-L/16 ImageNet &
  \multicolumn{1}{c}{-} &
  \multicolumn{1}{c}{-} &
  307M &
  0.852 &
  0.796 &
  0.847 &
  0.924 &
  0.669 &
  0.671 &
  0.793 \\
Lunit &
  21k &
  33M &
  22M &
  0.918 &
  0.771 &
  0.949 &
  \uline{0.943} &
  0.745 &
  0.756 &
  0.847 \\
CTransPath &
  32k &
  15M &
  28M &
  0.872 &
  0.817 &
  0.840 &
  0.930 &
  0.694 &
  0.726 &
  0.813 \\
Phikon &
  6k &
  43M &
  86M &
  0.906 &
  0.795 &
  0.883 &
  \textbf{0.946} &
  0.733 &
  0.751 &
  0.836 \\
Virchow &
  1.5M &
  2B &
  632M &
  0.933 &
  \textbf{0.834} &
  \uline{0.968} &
  - &
  - &
  - &
  - \\
RudolfV &
  103k &
  759M &
  307M &
  \uline{0.944} &
  0.821 &
  \textbf{0.973} &
  \uline{0.943} &
  \uline{0.755} &
  \uline{0.788} &
  \textbf{0.871} \\
GigaPath (patch encoder) &
  171k &
  1.3B &
  1,100 M &
  \textbf{0.947} &
  \uline{0.822} &
  0.964 &
  0.938 &
  0.753 &
  0.748 &
  \uline{0.862} \\
\name{} (ours) &
  35k &
  285M &
  86M &
  0.901 &
  0.818 &
  0.946 &
  0.939 &
  \textbf{0.756} &
  \textbf{0.804} &
  0.861 \\ \bottomrule
\end{tabular}%
}
\end{table}

Table \ref{tab:linear eval result} presents the classification accuracy results obtained from the linear evaluation. 
\name{} demonstrated competitive performance, achieving an average accuracy of 0.861 across the six downstream tasks. 
Notably, \name{} achieves the highest performance among all models on the MSI CRC dataset, with an accuracy of 0.756, and on the MSI STAD dataset, with an accuracy of 0.804.
Figure \ref{fig:param_comparison} compares the number of model parameters with the average accuracy across the six downstream tasks. 
Models positioned in the top left have fewer parameters and higher accuracy, indicating better efficiency. 
\name{} exhibits parameter efficiency compared to other models.
Figure \ref{fig:wsis_comparison} shows the average accuracy across six downstream tasks in relation to the number of WSIs used for model training. 
Models in the top left exhibit higher accuracy with fewer WSIs used for training. 
Training with stain-normalized images effectively mitigates the WSI-specific feature collapse problem, allowing for strong performance with fewer data and relatively smaller models.

\section{Conclusion}
In this paper, we have made a notable discovery during the visualization of features from patches extracted from models trained with self-supervised learning.
We have termed this phenomenon `WSI-specific feature collapse', where features from patches extracted from the same Whole Slide Image (WSI) tend to cluster together.
This suggests inefficient learning, as it leads to the extraction of similar features when thousands of patches are derived from a single WSI.
We hypothesized that this phenomenon arose from the substantial color variations between different WSIs.
To address this issue, we utilized Macenko normalization on the training data to minimize color differences and developed \name{}.
As a result, \name{} significantly reduced the WSI-specific feature collapse and achieved comparable or superior performance to existing models, even though those models use more data and parameters.
However, despite the considerable improvement in mitigating WSI-specific feature collapse through \name{}, this phenomenon appears to persist to some extent.
This suggests the need for further research. 
Specifically, developing more effective and computationally efficient stain normalization techniques, suitable for integration into data loaders, could be a promising direction for future research.
Additionally, exploring new learning methods or model architectures that can more effectively prevent feature collapse is also necessary.
In conclusion, \name{} represents significant progress in WSI analysis, laying the groundwork for more efficient and accurate pathological image analysis.

\bibliography{8.reference}

\newpage
\section*{EXAONEPath AI Model License Agreement 1.0 - NC}
This License Agreement (“Agreement”) is entered into between you (“Licensee”) and LG Management Development Institute Co., Ltd. (“Licensor”), governing the use of the EXAONEPath AI Model (“Model”). By downloading, installing, copying, or using the Model, you agree to comply with and be bound by the terms of this Agreement. If you do not agree to all the terms, you must not download, install, copy, or use the Model. This Agreement constitutes a binding legal agreement between the Licensee and Licensor.
\section*{1. Definitions}
\subsection*{1.1 Model}The artificial intelligence model provided by Licensor, which includes any software, algorithms, machine learning models, or related components supplied by Licensor. This definition extends to encompass all updates, enhancements, improvements, bug fixes, patches, or other modifications that may be provided by Licensor from time to time, whether automatically or manually implemented.
\subsection*{1.2 Derivatives}Any modifications, alterations, enhancements, improvements, adaptations, or derivative works of the Model created by Licensee or any third party. This includes changes made to the Model's architecture, parameters, data processing methods, or any other aspect of the Model that results in a modification of its functionality or output.
\subsection*{1.3 Output} Any data, results, content, predictions, analyses, insights, or other materials generated by the Model or Derivatives, regardless of whether they are in their original form or have been further processed or modified by the Licensee. This includes, but is not limited to, textual or numerical produced directly or indirectly through the use of the Model.
\subsection*{1.4 Licensor} LG Management Development Institute Co., Ltd., the owner, developer, and provider of the EXAONEPath AI Model. The Licensor holds all rights, title, and interest in the Model and is responsible for granting licenses to use the Model under the terms specified in this Agreement.
\subsection*{1.5 Licensee} The individual, organization, corporation, academic institution, government agency, or other entity using or intending to use the Model under the terms and conditions of this Agreement. The Licensee is responsible for ensuring compliance with the Agreement by all authorized users who access or utilize the Model on behalf of the Licensee.
\section*{2. License Grant}
\subsection*{2.1 Grant of License} Subject to the terms and conditions outlined in this Agreement, the Licensor hereby grants the Licensee a limited, non-exclusive, non-transferable, worldwide, and revocable license to:
\begin{enumerate}[label=\alph*.]
    \item Access, download, install, and use the Model solely for research purposes. This includes evaluation, testing, academic research and experimentation.
    \item Publicly disclose research results and findings derived from the use of the Model or Derivatives, including publishing papers or presentations.
    \item Modify the Model and create Derivatives based on the Model, provided that such modifications and Derivatives are used exclusively for research purposes. The Licensee may conduct experiments, perform analyses, and apply custom modifications to the Model to explore its capabilities and performance under various scenarios. If the Model is modified, the modified Model must include "EXAONEPath" at the beginning of its name.
    \item Distribute the Model and Derivatives in each case with a copy of this Agreement.
\end{enumerate}
\subsection*{2.2 Scope of License} The license granted herein does not authorize the Licensee to use the Model for any purpose not explicitly permitted under this Agreement. Any use beyond the scope of this license, including any commercial application or external distribution, is strictly prohibited unless explicitly agreed upon in writing by the Licensor.
\section*{3. Restrictions}
\subsection*{3.1 Commercial Use} The Licensee is expressly prohibited from using the Model, Derivatives, or Output for any commercial purposes, including but not limited to, developing or deploying products, services, or applications that generate revenue, whether directly or indirectly. Any commercial exploitation of the Model or its derivatives requires a separate commercial license agreement with the Licensor. Furthermore, the Licensee shall not use the Model, Derivatives or Output to develop or improve other models, except for research purposes, which is explicitly permitted.
\subsection*{3.2 Reverse Engineering} The Licensee shall not decompile, disassemble, reverse engineer, or attempt to derive the source code, underlying ideas, algorithms, or structure of the Model, except to the extent that such activities are expressly permitted by applicable law. Any attempt to bypass or circumvent technological protection measures applied to the Model is strictly prohibited.
\subsection*{3.3 Unlawful Use} The Licensee shall not use the Model and Derivatives for any illegal, fraudulent, or unauthorized activities, nor for any purpose that violates applicable laws or regulations. This includes but is not limited to the creation, distribution, or dissemination of malicious, deceptive, or unlawful content.
\subsection*{3.4 Ethical Use} The Licensee shall ensure that the Model or Derivatives is used in an ethical and responsible manner, adhering to the following guidelines:
\begin{enumerate}[label=\alph*.]
    \item The Model and Derivatives shall not be used to generate, propagate, or amplify false, misleading, or harmful information, including fake news, misinformation, or disinformation.
    \item The Model and Derivatives shall not be employed to create, distribute, or promote content that is discriminatory, harassing, defamatory, abusive, or otherwise offensive to individuals or groups based on race, gender, sexual orientation, religion, nationality, or other protected characteristics.
    \item The Model and Derivatives shall not infringe on the rights of others, including intellectual property rights, privacy rights, or any other rights recognized by law. The Licensee shall obtain all necessary permissions and consents before using the Model and Derivatives in a manner that may impact the rights of third parties.
    \item The Model and Derivatives shall not be used in a way that causes harm, whether physical, mental, emotional, or financial, to individuals, organizations, or communities. The Licensee shall take all reasonable measures to prevent misuse or abuse of the Model and Derivatives that could result in harm or injury.
\end{enumerate}

\section*{4. Ownership}
\subsection*{4.1 Intellectual Property} All rights, title, and interest in and to the Model, including any modifications, Derivatives, and associated documentation, are and shall remain the exclusive property of the Licensor. The Licensee acknowledges that this Agreement does not transfer any ownership rights to the Licensee. All trademarks, service marks, and logos associated with the Model are the property of the Licensor.
\subsection*{4.2 Output} All output generated by the Model from Licensee Data ("Output") shall be the sole property of the Licensee. Licensor hereby waives any claim of ownership or intellectual property rights to the Output. Licensee is solely responsible for the legality, accuracy, quality, integrity, and use of the Output.
\subsection*{4.3 Attribution} In any publication or presentation of results obtained using the Model, the Licensee shall provide appropriate attribution to the Licensor, citing the Model's name and version, along with any relevant documentation or references specified by the Licensor.
\section*{5. No Warranty}
\subsection*{5.1 “As-Is” Basis} The Model, Derivatives, and Output are provided on an “as-is” and “as-available” basis, without any warranties or representations of any kind, whether express, implied, or statutory. The Licensor disclaims all warranties, including but not limited to, implied warranties of merchantability, fitness for a particular purpose, accuracy, reliability, non-infringement, or any warranty arising from the course of dealing or usage of trade.
\subsection*{5.2 Performance and Reliability} The Licensor does not warrant or guarantee that the Model, Derivatives or Output will meet the Licensee’s requirements, that the operation of the Model, Derivatives or Output will be uninterrupted or error-free, or that defects in the Model will be corrected. The Licensee acknowledges that the use of the Model, Derivatives or Output is at its own risk and that the Model, Derivatives or Output may contain bugs, errors, or other limitations.
\subsection*{5.3 No Endorsement} The Licensor does not endorse, approve, or certify any results, conclusions, or recommendations derived from the use of the Model. The Licensee is solely responsible for evaluating the accuracy, reliability, and suitability of the Model for its intended purposes.
\section*{6. Limitation of Liability}
\subsection*{6.1 No Liability for Damages} To the fullest extent permitted by applicable law, in no event shall the Licensor be liable for any special, incidental, indirect, consequential, exemplary, or punitive damages, including but not limited to, damages for loss of business profits, business interruption, loss of business information, loss of data, or any other pecuniary or non-pecuniary loss arising out of or in connection with the use or inability to use the Model, Derivatives or any Output, even if the Licensor has been advised of the possibility of such damages.
\subsection*{6.2 Indemnification} The Licensee agrees to indemnify, defend, and hold harmless the Licensor, its affiliates, officers, directors, employees, and agents from and against any claims, liabilities, damages, losses, costs, or expenses (including reasonable attorneys' fees) arising out of or related to the Licensee's use of the Model, any Derivatives, or any Output, including any violation of this Agreement or applicable laws. This includes, but is not limited to, ensuring compliance with copyright laws, privacy regulations, defamation laws, and any other applicable legal or regulatory requirements.
\section*{7. Termination}
\subsection*{7.1 Termination by Licensor} The Licensor reserves the right to terminate this Agreement and revoke the Licensee’s rights to use the Model at any time, with or without cause, and without prior notice if the Licensee breaches any of the terms or conditions of this Agreement. Termination shall be effective immediately upon notice.
\subsection*{7.2 Effect of Termination} Upon termination of this Agreement, the Licensee must immediately cease all use of the Model, Derivatives, and Output and destroy all copies of the Model, Derivatives, and Output in its possession or control, including any backup or archival copies. The Licensee shall certify in writing to the Licensor that such destruction has been completed.
\subsection*{7.3 Survival} The provisions of this Agreement that by their nature should survive termination, including but not limited to, Sections 4 (Ownership), 5 (No Warranty), 6 (Limitation of Liability), and this Section 7 (Termination), shall continue to apply after termination.
\section*{8. Governing Law}
\subsection*{8.1 Governing Law} This Agreement shall be governed by and construed in accordance with the laws of the Republic of Korea, without regard to its conflict of laws principles.
\subsection*{8.2 Arbitration} Any disputes, controversies, or claims arising out of or relating to this Agreement, including its existence, validity, interpretation, performance, breach, or termination, shall be referred to and finally resolved by arbitration administered by the Korean Commercial Arbitration Board (KCAB) in accordance with the International Arbitration Rules of the Korean Commercial Arbitration Board in force at the time of the commencement of the arbitration. The seat of arbitration shall be Seoul, Republic of Korea. The tribunal shall consist of one arbitrator. The language of the arbitration shall be English.
\section*{9. Alterations}
\subsection*{9.1 Modifications} The Licensor reserves the right to modify or amend this Agreement at any time, in its sole discretion. Any modifications will be effective upon posting the updated Agreement on the Licensor’s website or through other means of communication. The Licensee is responsible for reviewing the Agreement periodically for changes. Continued use of the Model after any modifications have been made constitutes acceptance of the revised Agreement.
\subsection*{9.2 Entire Agreement} This Agreement constitutes the entire agreement between the Licensee and Licensor concerning the subject matter hereof and supersedes all prior or contemporaneous oral or written agreements, representations, or understandings. Any terms or conditions of any purchase order or other document submitted by the Licensee in connection with the Model that are in addition to, different from, or inconsistent with the terms and conditions of this Agreement are not binding on the Licensor and are void.\\

By downloading, installing, or using the EXAONEPath AI Model, the Licensee acknowledges that it has read, understood, and agrees to be bound by the terms and conditions of this Agreement.

\end{document}